Tech Science Press

# Roughsets-based Approach for Predicting Battery Life in IoT

**Rajesh Kaluri[1], Dharmendra Singh Rajput[1], Qin Xin[2,\*], Kuruva Lakshmanna[1], Sweta Bhattacharya[1], Thippa Reddy Gadekallu[1] and Praveen Kumar Reddy Maddikunta[1]**

[1]Vellore Institute of Technology, Vellore, Tamil Nadu, 632014, India
[2]Faculty of Science and Technology, University of the Faroe Islands, Vestarabryggja 15, FO 100, Torshavn, Faroe Islands
\*Corresponding Author: Qin Xin. Email: qinx@setur.fo

**Abstract:** Internet of Things (IoT) and related applications have successfully contributed towards enhancing the value of life in this planet. The advanced wireless sensor networks and its revolutionary computational capabilities have enabled various IoT applications become the next frontier, touching almost all domains of life. With this enormous progress, energy optimization has also become a primary concern with the need to attend to green technologies. The present study focuses on the predictions pertinent to the sustainability of battery life in IoT frameworks in the marine environment. The data used is a publicly available dataset collected from the Chicago district beach water. Firstly, the missing values in the data are replaced with the attribute mean. Later, one-hot encoding technique is applied for achieving data homogeneity followed by the standard scalar technique to normalize the data. Then, rough set theory is used for feature extraction, and the resultant data is fed into a Deep Neural Network (DNN) model for the optimized prediction results. The proposed model is then compared with the state of the art machine learning models and the results justify its superiority on the basis of performance metrics such as Mean Squared Error, Mean Absolute Error, Root Mean Squared Error, and Test Variance Score.

## 1 Introduction

Internet-of-Things (IoT) and related applications have gained immense momentum since the past decade with its implementations touching almost all spheres of life in our planet. IoT is often termed as Internet of Everything (IoE) which creates a global network of machines and devices integrated with one another establishing seamless communication systems across wide spectrum of domains [1,2]. With the rapid growth in human population and with surging usage of devices, services and related protocols the applications of IoT are numerous and hence become impossible to enlist [3]. The popular and extremely successful applications of IoT have been visualized in healthcare, agriculture, environment, automobile, transportation and defence [4,5].

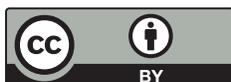

In healthcare, IoT applications have enabled reliable tracking of patients and transforming of data for providing essential and accelerated medical service [6]. The use of sensors has helped collect crop data thereby aiding agronomic decision making and enhancement of cultivation [7]. Industry 4.0 is the present trend for automation and data exchange in the manufacturing sector using cutting edge technologies like cyber physical systems (CPS), IoT, AI, cloud computing and Industrial IoT (IIoT). Industry 4.0 is a subset of the fourth generation industrial revolution which includes areas that are generally outliers in industrial sector. It basically encompasses a framework consisting of wireless connectivity, sensors and systems capable of making intelligent decisions providing visualization of the working of an entire manufacturing floor [8]. This concept sets the stage for smart factories. It is a giant progression from automation to a system which collects data streams from various operational processes and takes intelligent decisions for its running without human intervention. The mundane routine tasks are taken care by AI and machine learning applications making time for human beings to focus on only high level decision making and exception handling [9,10]. IoT environment have significantly contributed in traffic control, parking allotments and even assisted in improving driving experiences ensuring efficiency and safety [11,12]. Most importantly in the defense sector IoT applications have been adopted for surveillance, disaster prevention and robotic interventions [13]. The applications of IoT in agriculture, industry and various real time sectors have helped primarily in monitoring and control of weather conditions, health conditions, mechanical failures and various other instances of failure predictions. For all of these implementations sustainability of battery life is extremely important because if battery drains out it may lead to disastrous consequences due to failure in predictions of exceptional events [14–16].

The present study focuses on prediction of battery life sustainability for sensors which are installed in the beaches for collecting marine data. These sensors and batteries are primary components for IoT applications deployed to collect and monitor data from the marine environment namely Beach name, Measurement Timestamp, Water Temperature, Turbidity, Transducer Depth, Wave Height, Wave period and the battery life. Various applications for monitoring marine environment have been developed integrating IoT architectures, sensing, and control and communication technologies. These implementations are predominant in the areas of ocean sensing and monitoring, wave and current monitoring, water quality assessment, fish farming and coral reef monitoring. Although significant research has been conducted on IoT applications catering to the needs of marine environment, most of the studies have been conducted either at the under-water or surface level. Also solar energy harvesting has been a major point of discussion among various studies along with other forms of energy – wind, wave and ocean currents. The world at this point is progressing towards green technologies where energy optimization plays a vital role [17,18]. Based on such thorough analysis of the various applications of IoT in marine and other sectors, it has been observed that not much research have been conducted focusing on predicting the sustainability of battery life. Energy optimization is an essential aspect to be focused upon where the numerous research is being conducted emphasizing green computing. Naturally predicting the sustainability of battery life acted as a prime motivation in the present study.

The data used in the study is a publicly available dataset collected from the Chicago park district beach water. The missing values in the raw data are replaced with attribute means and then one hot encoding technique is applied achieve data homogeneity transforming all categorical values to numerical ones. This data is further normalized using Standard scalar technique and rough set approach is used for feature extraction. The extracted data thus contain best features having most significant impact on the target class – battery life. The extracted data is fed into a DNN ensuring use of activation function for optimized prediction results on battery life sustainability of the devices in the IoT framework. The prediction of battery life would help in tuning the connected sensor and network devices for the achievement of sustainable energy. The results are compared with the start of the art technologies to establish reliability of the prediction results.

The contributions of the proposed model are:

1. Using the DNN's a prediction model is proposed to forecast the battery life of the water sensors deployed along Chicago District Beach.
2. The proposed architecture uses the concepts of rough sets to minimize dimensionality and select significant features to increase prediction accuracy and decrease complexity.
3. The results are compared with common state-of-the-art techniques such as Linear Regression and XGBoost, which highlight the enhanced performance of the proposed model
4. A review of the current rough sets model and the state-of-the-art techniques is performed in depth.
5. The experimentation justify the fact that the proposed prediction model outperforms other popular ML techniques thereby validating the positive impact of integrating rough set technique with the ML model.

## 2 Literature Review

In the recent decade numerous artificial intelligence methods have been used to boost the learning level of the systems in an extremely granular manner. In this section, discussion on the various contributions of deep learning algorithms and rough set concepts are discussed through an extensive survey of literature.

In [19] authors have developed the process by which deep learning and rough sets approaches can automate the evaluation network for motors that work through highways in snow and ice environments. The experiment was conducted first, on a motorway in Jilin, China, with a naturalistic driving test involving 13 licensed drivers, with specific crash point set before the inception of the experiment. It was used to collect images and numerical data with multi-sensors (eye trackers, mini camera, and speed detectors). The Boltzmann Restricted machine was used for the development and training of a deep-belief network (DBN). The technique of Rough Sets was added as an evaluation of the DBN output layer. Various conditions for perception of input factors were used to train the network. The findings emphasize that DBN-FS was not only higher in the Naive Bayes and the BP-ANN but also greater precision in sensing dangerous conditions due to the comparative implementation with Naive Bayes and BP-ANN of the DBN-based perception network. This approach was useful to design partially automated vehicle hazard detection systems.

In [20] authors proposed a DNN hybrid architecture, stacked autoencoder (SAE) and stacked denoising autoencoder (SDAE) for ultrashort-term and short-term wind velocity prediction. Auto-encoders (AEs) were used to learn unsupervised feature learning from a supervised regression layer and wind data was used for wind speed forecasting. It was observed that many unknown factors had significant effect on the accuracy of existing methodologies used in prediction of wind velocities. Deep learning models with RNN have been used to develop new rough extensions of SAE and SDAE that stood to wind uncertainties and ensure higher accuracy in forecasting. Experimental results of this study proclaimed the proposed rough DNN models are superior to the traditional DNNs and various existing methods which apply shallow architectures and mean absolute error measurements.

In [21] authors provided a novel aggregation algorithm for wireless sensor networks using the combination of rough set concepts with an improvised version of Convolutional Neural Network (CNN). The designed model was used for the extraction of features and training in the sink node. The rough set concept was adopted in order to simplify information effectively and to reduce tagged dimensions. When data features were removed from the granular deep network by cluster nodes, these data functionalities have been transferred to the sink node to further increase the life span of the network by reducing data density. Simulation results in the study helped to compare the existing data aggregation algorithms wherein in terms of energy consumption, the proposed granular CNN model contributed significantly towards enhancement of accuracy in data aggregation.

In [22] authors provided a brief summary and functionality of the association of IoT with sensing and wireless technologies for the implementation of the necessary health applications. In [23] authors proposed a prediction model based on DNN, which used as a patent indicator as predictors by the company's financial. Proposed approach has two phases: a learning phase and another is a fine-tuning phase. The learning stage uses limited Boltzmann machines and the fine-tuning stage uses the latest data sets to implement a back propagation algorithm, which reflect the recent trends pertinent to relationships between predictor-company relations.

In [24] authors checked the performance of generated data using the IEEE 802.15.4 standard. The DNN was used to model the relationship between different communication parameters and delays. The evaluations identified DNN model being capable of achieving optimum level of forecast accuracy outperforming other popular regression models. The study also highlighted the success of the model even further due to its ability to generate comparable accuracy when trained with small fraction of data.

In [25] authors focused on development and implementation of multi-source data fusion algorithmic model (JDL) for the diagnosis and prediction of faults in mechanical equipment. The JDL fusion model was relatively mature, yet best suited for mechanical fault diagnostic applications. The ideas of this approach correlated with that of the hierarchical fusion model and hence were prevalent in the diagnosis of mechanical faults. The basis of this model was neural network in convergence with data pre-processing algorithms like SVD, EMD and WPD to achieve better accuracy. Similarly fusion algorithms were deployed based on deep learning approaches capable of directly training and extracting data for analysis and prediction purposes.

The management of on-site maintenance visits is an important requirement for the successful operation of IoT networks. In particular, on-site visits to replace or recharge depleted batteries may be required in battery powered devices. For this purpose, practitioners often use prediction techniques to estimate the battery life of the deployed devices. Battery life is one of the biggest challenges for IoT at present [26].

In [26] authors concentrate on the actual battery lifetime and battery lifetime predictions during the development of the deployed system with discharge patterns. The result shows the difficulties of making predictions about battery life in an effort to encourage more study. In [27] authors have used four different machine learning techniques that are used to handle the problem of remaining useful life battery estimates under different flight conditions. Based on multiple experimental flight data conditions, the effectiveness of the overall machine learning techniques in the field of battery prognostics are evaluated.

In [28] authors present a deep learning method that employs DCNN to estimate cellular capacity based on measurements of voltage, current, and charging capacity for a partial charge cycle. In accordance with our knowledge, this is one of the first attempts to apply a deep learning to online Li-ion battery capacity assessment.

Tab. 1 present a comparative study of the various research works conducted in the field of Deep Learning and Roughsets theory.

## 3 Preliminaries

In this section the algorithms and methods used in this study are explicitly discussed.

### 3.1 Rough Sets

The conventional mathematical methods like crisp sets failed to resolve unclear and vague problems which acted as the prime motivation among researchers to work in a direction to solve such problems where approximations would play a major role. The concept of Rough sets was first proposed in [29] where the primary emphasis was on selection and extraction of features, data reduction and generation of decision rule.

**Table 1:** Comparative study among different Deep Learning and Rough Sets

| Author | Contribution | Advantages | Disadvantages |
|---|---|---|---|
| Lopez et al. [30] | A classifier for monitoring Network traffic is designed hybridizing RNN with CNN. | Robust Method and F1 detection scores is excellent. | High-level header based data is used extracting from the data packets for training the model. |
| Fan et al. [31] | PSO algorithm and Rough PSO algorithm is proposed considering the rough set Concept | Rough PSO also is used for classification. Faster convergence speed also better precision | Trapped in locally optimum solution and Time complexity |
| Mahdavinejad et al. [32] | ML Techniques are studied. Applications of IoT, IoT data characteristics and relevant technical implementations are explored. | The methods surveys have been identified to be suitable with ease of use in various types of problems | Present issues in the field of smart data analytics have not been included. |
| Manogaran et al. [33] | An optimized Bayesian neutral network is proposed consisting of densely connected layer is proposed. The model helps to detect temperature imbalance in the field of healthcare. | Problems relevant to multiple access based physical monitoring system is resolved. | Methods to scale the devices is excluded from the study. |
| Chakraborty et al. [34] | Make use of Neighborhood Rough Set (NRS) in exploiting the uncertainty of tracking the object in a video sequence. | Without any prior knowledge and variations in size and speed also, propose the rack of multiple objects. | In the areas of signal processing, the main concern of NRS filter and intuitionistic entropy are unsupervised prediction and handling ambiguity. |
| Hassan et al. [35] | A novel approach is proposed based on concepts of rough sets to develop ML and soft computing using DL architecture. The method caters to solve real-life problems. | Approach focuses on integration of local properties pertinent to individual decision table. The method attempts enables acceptable global decision making. | RS assumes the presence of single decision-making table, whereas real-world problems involve numerous decisions from varied decision-making tables. |
| Otero et al. [36] | ACO algorithm is proposed incorporating widely used techniques from standard decision tree and ACO algorithms. | Comparison of 22 publicly available data sets including three decision tree algorithms - CACDT, CART, and C4.5 | Results are statistically proof but with single test has done. |

Fuzzy and rough sets have been considered as complimentary simplifications of the classical set theory concepts. The recourse to two definable subsets (ψ) is loosely defined by subsets of a universal structure (U).

As per the theory of Rough sets, there exist four classes namely -

A is roughly X - definable, iff X(A)≠ ψ ∧ X⁻ (A) ≠U

A is roughly X - definable, iff X(A) = ψ ∧ X⁻ (A) ≠U

A is roughly X - definable, iff X(A) ≠ ψ ∧ X⁻ (A) = U

A is roughly X - definable, iff X(A) = ψ ∧ X⁻ (A) = U

### 3.2 Accuracy

Accuracy of any rough set (A) can be measured by the possible values of target set (A) to the immediate set.

$$\alpha X(A) = |XA|/|XA^-| \qquad (1)$$

where A - represents the carnality of set A and α X(A) lies in between 0 and 1.

### 3.3 Attribute Dependency

It is the most significant element in determining the variables from the rough sets. Considering two sets (X, Y), its equivalence classes is considered as [A]X, [A]Y where $Y_i$ is the equivalence class from the attribute set Y. The dependency for the other set X, η can be calculated as

$$\eta = \gamma(X, Y) = \sum_{i=1}^{n} \frac{|A(X)Y_i|}{U} \qquad (2)$$

From Eq. (2), it is evident that If γ(X, Y) = 1 then Y depends on X, on the contrary, if γ(X, Y) ≤ 1 then Y partially depends on X.

### 3.4 Deep Neural Network

DNN is one of the standard methods for generating classification models where learning can be of any forms supervised, unsupervised or semi-supervised [37,38]. DNN parameters are used to examine recognition tasks, which tend to provide best performance over datasets. A DNN builds a deep architecture by taking self-encoders for the representation of hierarchical features. The Fig. 1 illustrates the structure of DNN architecture, consisting of an input layer, hidden layers, and an output layer. The data set acts as an input to the input layer.

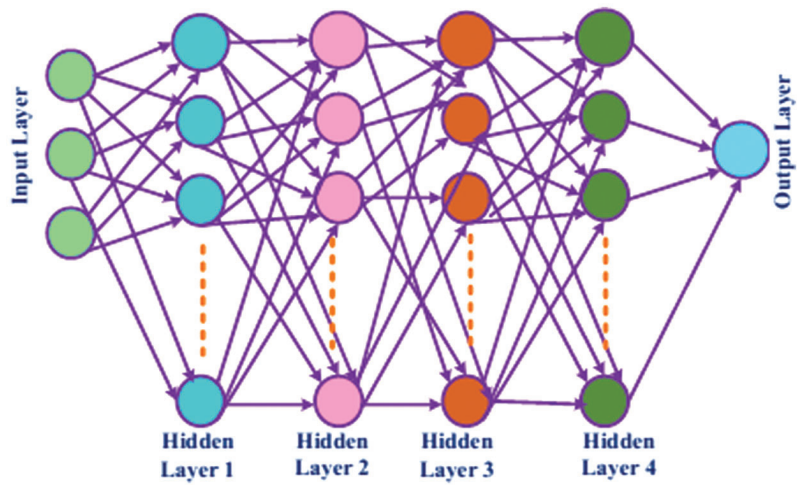

**Figure 1:** Deep Neural Network (DNN)

The input to neuron in the first hidden layer is given in Eq. (3)

$$A_{1(x)} = \sum C1\ (x, y)\ ay\ +\ z1\ (x) \tag{3}$$

where $C_{1\ (x,y)}$ and $z_{1\ (x)}$ are the weight and bias respectively. The output of the neuron x in the first hidden layer is given by $h_{1(x)} = $ *Activation Function* $(A_{1(1)})$ where *Activation Function* (.) is a particular activation function used for the DNN.

### 3.5 *XGBoost*

XGBoost is one of the most effective approaches in ML which implements the concept of decision trees in regular successions without any gaps [39]. Considering the aspect of computing paradigms, the model primarily uses gradient boosting framework for optimized estimations. Fig. 2 gives a detailed view on the various forms of gradient boosting and the process of effective data handling using paralleled tree building segment. XGBoost basically is a form of boosting algorithm that convert weak learners into stronger ones which basically improves the process of random guessing. Boosting as a popular methodology is a sequential process wherein trees are developed using information of previously developed trees in a sequential process following one another. Hence by learning from data, the process improves the resultant predictions of the following iterations. The algorithm also renders good rate of regularization in boosting the model and handles the missing values appropriately. Hence, XGBoost stands extremely appropriate for classifying the labels by modeling its attributes. Numerous researchers have prioritized the use of XGBoost algorithm in comparison to other algorithms due to its ability to generate accurate results ensuring optimal performance level.

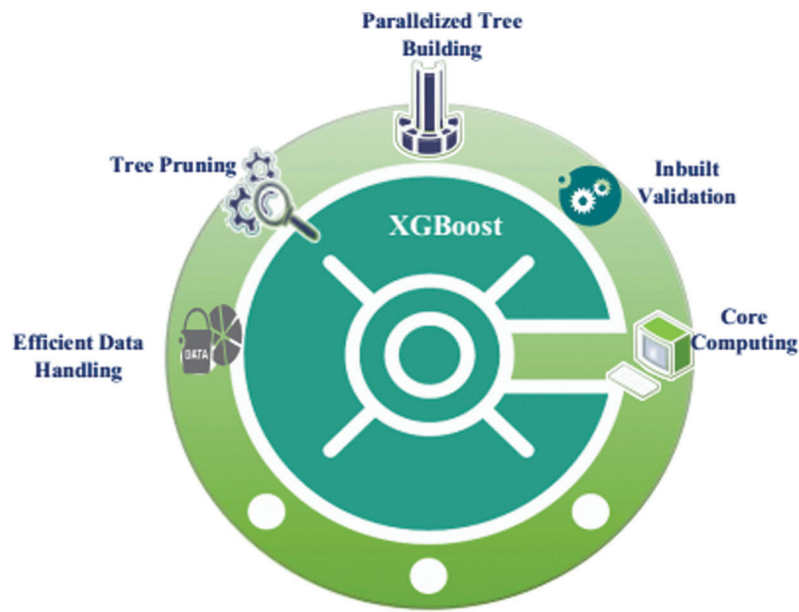

**Figure 2:** XGBoost

### 3.6 *Linear Regression*

Linear Regression is one of the best supervised algorithms with declared output having perpetual and constant slope. It predicts values from a range of enclosed data values. Simple regression and multi-variable regression are the forms of Linear Regression. Simple regression and multi-variable regression are usually referred by

$$y = mx + C \tag{4}$$

$$f(x, y, z) = a_1 + a_2 y + a_3 z \tag{5}$$

where x,y,z represents attributes, m, $a_1$, $a_2$ and $a_3$ are variables for the regression process. Statistical methods can be used to measure and reduce the size of the error variable to improve predictive power of the model.

### 3.7 One Hot Encoding

One Hot Encoding is a process of exemplifying categorical values into binary numbers. It is a known fact that machine learning algorithms fail to work on categorical data and hence have to be converted to numbers where one hot encoding technique plays its significant role. As a natural reaction it is possible to opine on the use of integer coding directly but it has its limitations when used on relationships of the natural ordinal types. In one hot encoding technique, the categorical values in the data are directly assessed to integer values and each integer value is changed to a binary value [40].

## 4 Proposed Model

The architecture of the proposed model is depicted in the Fig. 3. The steps involved are: The IoT dataset collected from the Chicago District Beach Water District [41]. The sensors are used in six different district locations to monitor and detect indicative measurements of beach name, date of measurement, water temperature, turbidity, transducer depth, wavelength, and wave duration for six places per hour during the summer.

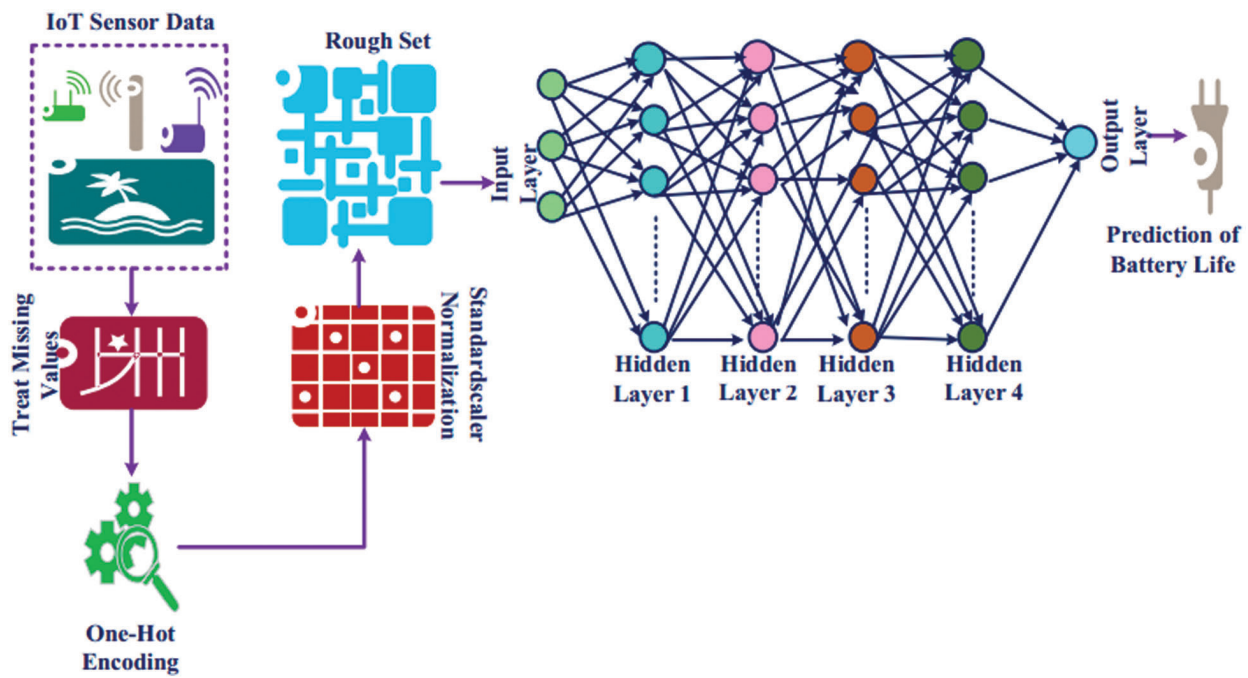

**Figure 3:** Proposed Rough Set Based DNN Architecture

a. Preprocessing plays a crucial role in machine learning algorithms.

- **Treat Missing Values:** The missing values in the dataset are filled with the respective attribute mean, which makes the dataset ready for further analysis. The dataset consists of ten features, out of which two features are time tag driven, which do not contribute to the battery life prediction of the sensor node. Therefore, these two features are omitted for further analysis. The remaining eight features consist of categorical data, which also requires processing.

- **One-Hot Encoding:** In the next step of processing, the one-hot encoding scheme converts the categorical data into numerical data. There are eight features with various instances that get extrapolated to 206 features.

- **Standard scaler Normalization:** Further, the standard scaler normalization technique is applied on the numerical dataset, which enables standardization of the features transforming the mean of the distribution to 0. This ensures most of the values in the dataset range within 0 and 1.

b. The dimensionality reduction technique is applied to the used data using rough Set theory.

c. The reduced dataset is then fed into a deep neural network model for predicting the sustainability of battery life.

d. The DNN model is evaluated against the traditional state of the art models to validate the performance of the predictions.

## 5 Results and Discussion

*Dataset Description:* The experimentation was carried out in a personal laptop with 8GB RAM, 500GB of Hard-disk, Windows 10 operating system and coding was done using Python 3.8. The dataset used for this experimentation is of the Chicago Park District Beach Water collected from [41]. There exists specific sensors mounted in the water at specific beaches along the Michigan Lakefront in Chicago Park District. These sensors are used in six locations in the district to track and detect indicative measurements pertaining to the name of the beach, the date of measurement, the temperature of the water, the turbidity, and the depth of the transducer, the wave length and the wave duration for six places per hour during the summer. The attributes of the dataset are described in the below Tab. 2. The dataset holds water sensor data for 10 characteristics in all 39.5 K cases. As shown in the following Tab. 3, the location where water sensors are mounted on the beaches.

**Table 2:** Chicago Park District Beach Dataset

| Column Name | Data Description | Unit of the Gathered Sensor Data | Range of Data |
|---|---|---|---|
| Beach Name | Name of the beach where water sensor is deployed | Plain Text String | 6 locations |
| Measurement Timestamp | Measurement is done on hourly basis in a single day throughout the season | Date and Time | 30/8/2013 8.00 AM & 11/9/2019 11.00 AM |
| Water Temperature | Temperature of water at the time of measurement | Degree Celsius | 9.1 to 31.5 |
| Turbidity | Turbidity of the water waves | Nephelometric Turbidity Units (NTU) | 0.01 to 1683.48 |
| Transducer Depth | Depth | Meters | −0.082 to 2.214 |
| Wave Height | Height of the water waves | Meters | 0.013 to 1.467 |
| Wave Period | Period at which the waves occur | Seconds | 1 to 10 |
| Battery Life | Voltage of the battery remaining for the purpose of deciding the time for replacement | Numeric | 4.8 to 13.3 |

**Table 3:** Location of Water Sensors Deployed in the Beaches

| Name of the Location of the water sensor | Location of the Sensor (in Latitude and Longitude) |
| --- | --- |
| 63rd Street Beach | (41.784561°, −87.571453°) |
| Calumet Beach | (41.714739°, −87.527356°) |
| Montrose Beach | (41.969094°, −87.638003°) |
| Ohio Street Beach | (41.894328°, −87.613083°) |
| Osterman Beach | (41.987675°, −87.651008°) |
| Rainbow Beach | (41.760147°, −87.550081°) |

### 5.1 Metrics used for Evaluating the Model

**Mean Squared Error (MSE):** This is basically an average estimator that allows the researchers to determine the difference between the value and the actual value.

$$MSE = \frac{1}{m} \sum_{i=1}^{m} (x_i - \tilde{x}_i)^2 \tag{6}$$

where $x_i$ orthoginal value and $\tilde{x}_i$ is the predicted value and m is the number of observations.

**Mean Absolute Error (MAE):** The mean absolute error metric measures the difference between a pair of continuous measurements in a single specific phenomenon. It reports that all the findings in progress show the same absolute error.

$$MAE = \frac{1}{m} \sum_{i=1}^{m} |x_i - \tilde{x}_i| \tag{7}$$

**Root Mean Squared Error (RMSE):** The statistics are commonly used to assess the magnitude of the prediction error and aggregates the same to a single measurement for assessing the strength of the prediction.

$$RMSE = \sqrt{\frac{1}{m} \sum_{i=1}^{m} (x_i - \tilde{x}_i)^2} \tag{8}$$

The MAE calculates similarity of the projections to potential results, while the RMSE is the standard deviation of the sample from the discrepancies between the expected $\tilde{x}_i$ and the observed $x_i$ values, which measure the intensity and direction of a linear $x_i - \tilde{x}_i$ relation.

**Test Variance Score (TVS):** This metric calculates the variance between the actual values and observed values from the cumulative average of all expected rates generated by the forecast model or algorithm. The main goal of the predictor is to have a low variance performance. h is the sum of the variations between the observed and the predicted values.

$$h = (ztest[i] - \tilde{x}_i)^2 \tag{9}$$

$$z = \frac{observed - actual}{standarderror} \tag{10}$$

z is every z[i] value that is observed minus the sum of the average of the ztest values that are observed. The variance is,

$$TVS = 1 - \frac{h}{z} \tag{11}$$

## 5.2 Dataset Preparation for Prediction

Real time raw data for water are collected from different places. Since the data is raw, there are few instances with some features missing, as the battery drain the sensor reducing battery sustainability. There is a possibility that some time-stamp instances would also not be usable until the battery is replaced. But filling in of missing values is extremely important before initiating the process of forecasting using the ML model. In the present study, the missing values in the dataset are first filled with the the respective attribute mean which makes the dataset ready for further analysis.

It is known from the data summary that out of ten characteristics, two are time tag driven, which do not contribute to the battery life prediction of the sensor node. Therefore, both features are omitted for further analysis. The remaining eight features of the Tab. 2 consist of categorical data which also requires processing. Thus, the one-hot encoding scheme converts these categorical data into numerical data. There are eight features with various instances that are extrapolated to 206 features. The transformed sensor results were then normalized from 0 to 1 using the Standardscalar method.

The next attempt was directed to extract features having higher impact on battery life prediction. Also the ones with negative impact were eliminated using rough sets. The number of features thus got decreased to 182 as part of dimensionality reduction. These features have the best impact on the target class prediction.

**Prediction Model:** The DNN used in the prediction model consisted of one input layer with 128 neurons, five hidden layers with 256, 256, 256, 128, and 64 neurons respectively and one output layer with 1 neuron. The Adaptive Moment Estimation (ADAM) Optimizer and the Relu activation function are used for compiling the prediction model. Tabs. 4 and 5 provide an overview of a DNN prediction process considering the use of rough sets in contrast with non-inclusion of the rough set technique.

**Table 4:** Prediction Model DNN with rough sets as Dimensionality Reduction

| Layer | Type | Shape | Parameters |
|---|---|---|---|
| Dense\_17 | Dense | 128 | 22123 |
| Dense\_18 | Dense | 256 | 31451 |
| Dense\_19 | Dense | 256 | 63652 |
| Dense\_20 | Dense | 256 | 63652 |
| Dense\_21 | Dense | 256 | 52244 |
| Dense\_22 | Dense | 128 | 31475 |
| Dense\_23 | Dense | 64 | 7856 |
| Dense\_24 | Dense | 1 | 61 |
| Total Parameters | 272514 | | |
| Trainable Parameters | 272514 | | |

**Performance Evaluation:** The proposed predictive model is evaluated against other commonly used state-of-the-art models, namely DNN, Linear Regression and XGBoost, for validating the predictive effect of the model using DNNs with rough sets integrated for dimensionality reduction. The results integrating rough sets with DNN and without are critically analyzed. The data is shown in the Figs. 4–6 highlights the effects of the random selection for 25 data instances. The same experiment is repeated after reducing the dimension using rough sets with respect to all the prediction models. The obtained results are shown in the Figs. 7–9. Post observation of the all the images, the proposed model shows the better results when we compare the other models like DNN, Linear Regression, XGBoost without roughsets.

**Table 5:** Prediction Model DNN without Dimensionality Reduction

| Layer | Type | Shape | Parameters |
|---|---|---|---|
| Dense\_14 | Dense | 128 | 25624 |
| Dense\_15 | Dense | 256 | 32123 |
| Dense\_16 | Dense | 256 | 64634 |
| Dense\_17 | Dense | 256 | 64634 |
| Dense\_18 | Dense | 256 | 64634 |
| Dense\_19 | Dense | 128 | 31684 |
| Dense\_20 | Dense | 64 | 7956 |
| Dense\_21 | Dense | 1 | 54 |
| Total Parameters | 291343 | | |
| Trainable Parameters | 291343 | | |

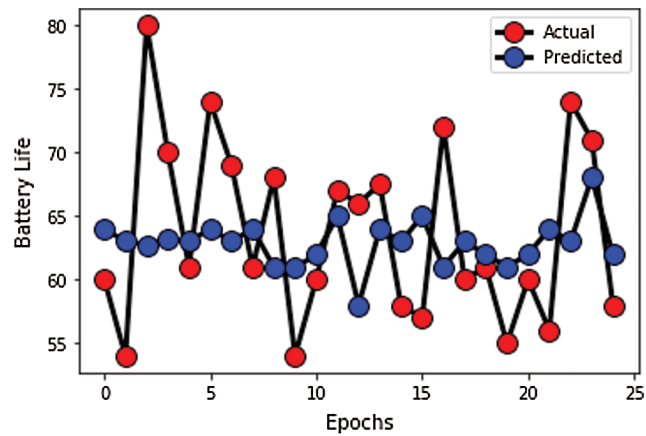

**Figure 4:** Linear Regression Model Plot

The proposed model is evaluated on the basis of various standard performance metrics: Mean Absolute Error, Mean Squared Error, Root Mean Squared Error, and Test Variance [42–47]. A comparison of different machine learning models is shown in Tab. 6, values of MAE indicate the superiority of Roughsets DNN where DNN, Roughsets Linear Regression, linear regression, Roughsets XGBoost and XGBoost achieve MAE values of 5.16, 5.25, 6.45, 979.62, 5.34 and 5.47, respectively.

The Figs. 4–9 highlight the residual battery life (mAh) for the various prediction models namely Linear Regression, XGBoost, DNN, Linear Regression + Rough Set, XGBoost + Rough Set, DNN + Rough Set. It is pertinent to mention that the primary focus of the work was the prediction of battery life in terms of residual life of battery when used in sensors of Chicago district beach water. The more life remains, the working ability would automatically get extended for data collection of these sensors for prolonged time period.

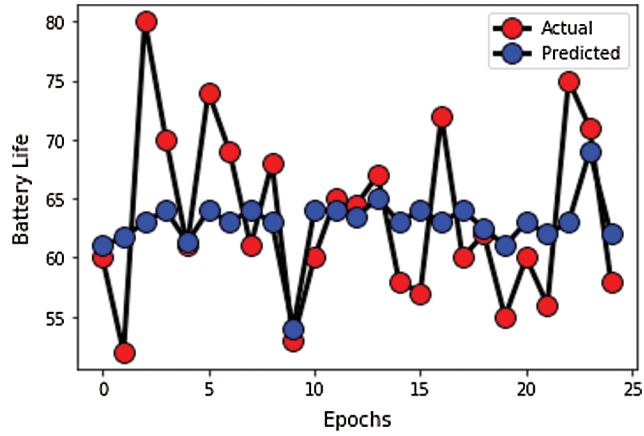

**Figure 5:** XGBoost Model Plot

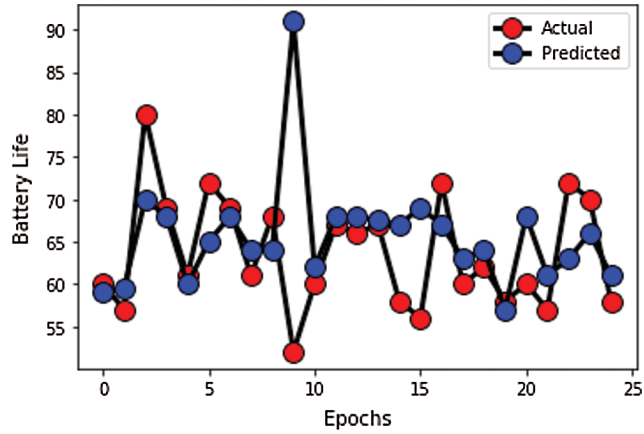

**Figure 6:** Deep Neural Networks Prediction Model

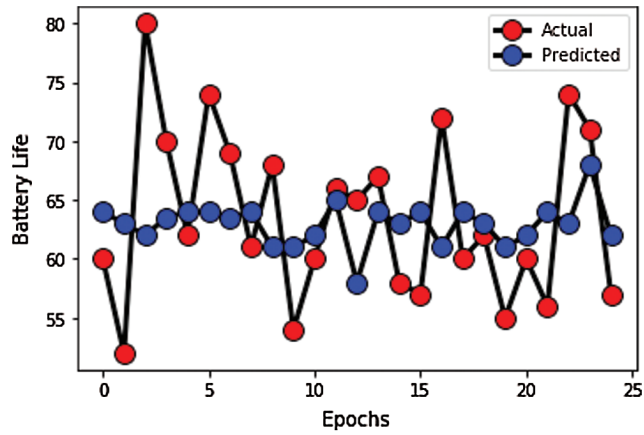

**Figure 7:** Linear Regression with Rough sets

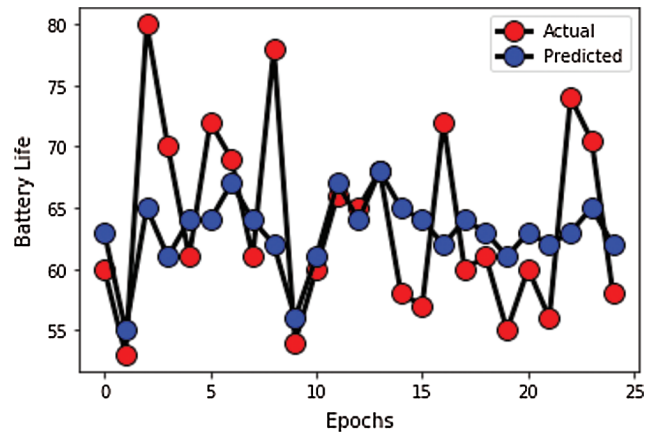

**Figure 8:** XGBoost with Rough sets

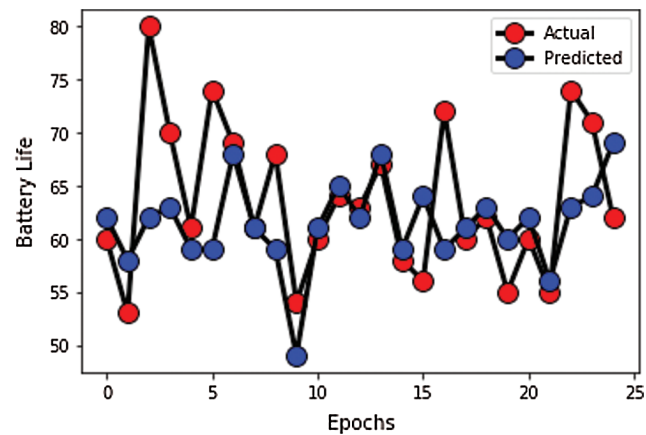

**Figure 9:** Deep Neural Networks with Rough sets

**Table 6:** Evaluation of Prediction Models using Standard Metrics

| Model Used | MAE | MSE | RMSE | TVS |
|---|---|---|---|---|
| DNN + Rough sets (Proposed) | 5.16 | 35.68 | 5.85 | 0.11 |
| DNN | 5.25 | 36.81 | 5.98 | 0.17 |
| Linear Regression + Rough sets | 6.45 | 80.47 | 8.74 | −0.23 |
| Linear Regression | 979.62 | 16920158.899 | 40786.5 | −256623.7 |
| XGBoost + Rough sets | 5.34 | 45.58 | 6.12 | 0.26 |
| XGBoost | 5.47 | 44.12 | 6.31 | 0.29 |

## 6 Conclusion and Future Work

There is no doubt that there are significant studies relevant to application of IoT in marine environment. But there exist lag in research which focus on energy optimization aspects of IoT applications especially towards enhancement of sustainability of battery life. In the present study a novel approach has been adopted based on converging rough set with DNN to predict battery life of the IoT network with optimum accuracy. The normal pre-processing steps used in this approach were further refined incorporating the rough set approach for extracting significant features which has contributed immensely towards more accurate predictions. The results of the model were compared with the state of the art techniques to establish its superiority. As part of future work, the same model can be deployed on several IoT applications rendering home surveillance, healthcare and defence services. Also the scalability and robustness of the model can be validated by testing the same on magnanimous IoT application like traffic predictions, air pollution, waste management etc.in smart city setups. The predictions of battery life would also guide in the design and development of energy efficient products involving IoT and sustainable energy technologies.

**Funding Statement:** The author(s) received no specific funding for this study.

**Conflicts of Interest:** The authors declare that they have no conflicts of interest to report regarding the present study.